\documentclass[runningheads,a4paper]{llncs}

\usepackage{amssymb}
\usepackage{graphicx}

\usepackage{url}
\urldef{\mailsa}\path|{alfred.hofmann, ursula.barth, ingrid.haas, frank.holzwarth,|
\urldef{\mailsb}\path|anna.kramer, leonie.kunz, christine.reiss, nicole.sator,|
\urldef{\mailsc}\path|erika.siebert-cole, peter.strasser, lncs}@springer.com|

\usepackage{epsfig}
\usepackage{amsmath}
\usepackage{amssymb}
\usepackage{booktabs}

\usepackage{caption}
\usepackage{subcaption}
\usepackage{breakcites}

\def\onedot{. }
\def\eg{\emph{e.g}\onedot} 
\def\ie{\emph{i.e}\onedot}

\def\etal{\emph{et al}\onedot}

\usepackage{tikz,xcolor,hyperref}

\definecolor{lime}{HTML}{A6CE39}
\DeclareRobustCommand{\orcidicon}{
	\begin{tikzpicture}
	\draw[lime, fill=lime] (0,0) 
	circle [radius=0.16] 
	node[white] {{\fontfamily{qag}\selectfont \tiny ID}};
	\draw[white, fill=white] (-0.0625,0.095) 
	circle [radius=0.007];
	\end{tikzpicture}
	\hspace{-2mm}
}

\foreach \x in {A, ..., Z}{\expandafter\xdef\csname orcid\x\endcsname{\noexpand\href{https://orcid.org/\csname orcidauthor\x\endcsname}
			{\noexpand\orcidicon}}
}

\begin{document}

\mainmatter  % start of an individual contribution

%\title{Learning Spectral-Specific Graph Representation for Hyperspectral Image Classification}
\title{Spectral Graph Reasoning Network for Hyperspectral Image Classification}

\titlerunning{Spectral Graph Reasoning Network for Hyperspectral Image Classification}

\author{Huiling Wang\orcidA{}\thanks{huiling.wang@tuni.fi}}
\authorrunning{Huiling Wang}

%\institute{Nokia Technologies, Finland
%	\and
%	Tampere University, Finland}
%\authorrunning{Tinghuai Wang, Huiling Wang}
%\author{%
%	Huiling Wang\inst
%}%
\institute{Computing Sciences, Tampere University, Finland}


%\institute{Springer-Verlag, Computer Science Editorial,\\
%Tiergartenstr. 17, 69121 Heidelberg, Germany\\
%\url{huiling.wang@tuni.fi}}


%\toctitle{Lecture Notes in Computer Science}
%\tocauthor{Authors' Instructions}
\maketitle

\begin{abstract}
	Convolutional neural networks (CNNs) have achieved remarkable performance in hyperspectral image (HSI) classification over the last few years. Despite the progress that has been made, 
	rich and informative spectral information of HSI has been largely underutilized by existing methods which employ convolutional kernels with limited size of receptive field in the spectral
	domain. To address this issue, we propose a spectral graph reasoning network (SGR) learning framework comprising two crucial modules: 1) a spectral decoupling module which unpacks and casts
	multiple spectral embeddings into a unified graph whose node corresponds to an individual spectral feature channel in the embedding space; the graph performs interpretable reasoning to aggregate and align spectral information to guide
	learning spectral-specific graph embeddings at multiple contextual levels 2) a spectral ensembling module explores the interactions and interdependencies across graph embedding hierarchy via a novel recurrent graph propagation mechanism. Experiments on two HSI datasets demonstrate that the proposed architecture can significantly improve the classification 
	accuracy compared with the existing methods with a sizable margin.
	
	%Existing CNN architectures are typically utilizing 
	%standard convolutional kernels with fixed weights given their position within the convolution window. This limitation of the standard convolutional kernel neglects 
	%the structural connections between data points, resulting in poor region delineation and small spurious predictions.
	
	%Existing CNN architectures are typically designed and optimized for classifying RGB images, which excel in capturing the 2D spatial contexts while learning a rich feature representation. However, HSIs 
	%normally have a unique nature of data distribution along the high dimentional spectrum domain - much remains to be addressed in capturing the  spectral contexts considering the prohibitively high dimensionality 
	%and improving reasoning capability in light of the limited amount of labelled data. 

\end{abstract}

\begin{figure*}[htbp]
	\centerline{\includegraphics[width=1.0\linewidth]{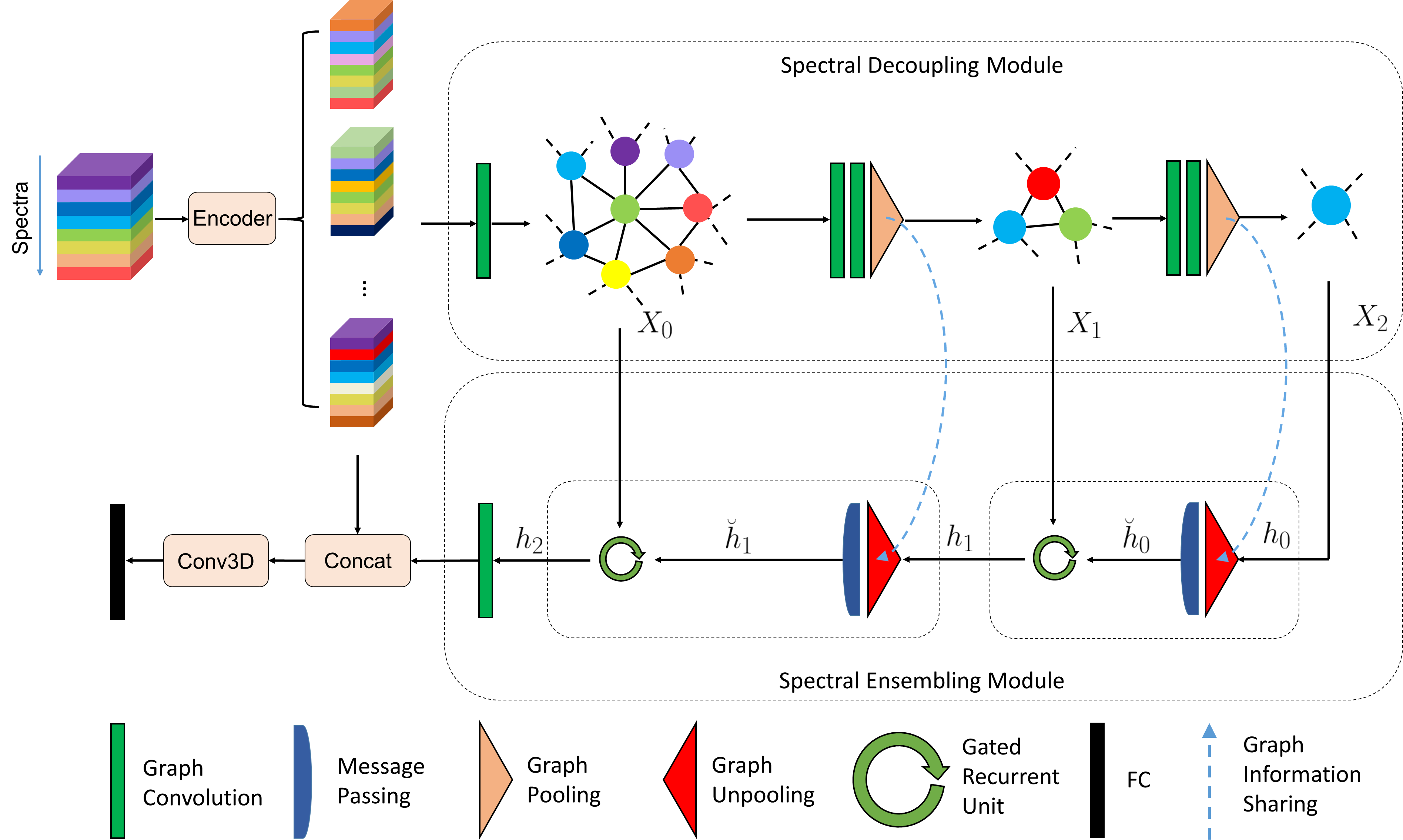}}
	\caption{Illustration of our proposed network architecture.}
	\label{diagram}
\end{figure*}

\section{Introduction}
Hyperspectral image comprises hundreds of continuous spectral bands throughout the electromagnetic spectrum with high spectral resolution, which facilities the precise differentiation of chemical properties of scene materials remotely. 
Consequently hyperspectral images have been considered as a crucial data source in various fields, such as environmental monitoring, mining, agriculture, and land-cover mapping. 

HSI classification involves assigning a categorical class label to each individual pixel location given the corresponding spectral feature. Various classification approaches have been proposed to address the hyperspectral image classification problem. 
Early approaches largely adopted traditional machine learning methods which were trained on hand-crafted features from HSI data to empirically encode the spectral information, \eg SVM \cite{kuo2010spatial}, KNN \cite{ma2010local}, dictionary learning \cite{chen2011hyperspectral}, graphical model \cite{shi2012supervised}, extreme learning machine \cite{li2015local} and among others. Other methods \cite{zhong2014discriminant,zhang2016simultaneous} also exploited both spectral and spatial information 
since utilizing only the spectral information regardless of the spatial correlation is difficult in accurately classifying different land-cover categories.

With the advent of new remote sensing instruments and increased temporal resolutions, the deluge of high dimensional HSI data is posing new challenges on the limited discriminative power of the empirically designed features  and the traditional classification methods. Deep learning based HSI classification methods have been proposed inspired by its success in visual recognition of natural images. Those methods benefit from the strong representation capability and end-to-end learning.  Chen \emph{et al.} \cite{chen2014deep} developed the first deep neural networks (DNN) method which utilized stacked autoencoders to learn high-level features. Mou \emph{et al.} \cite{mou2017deep} introduced a Recurrent Neural Network (RNN) based architecture. More powerful end-to-end Convolutional Neural Network (CNN) based architectures \cite{chen2016deep,hamida20183,hu2019spnet}  have been proposed recently. Lee \emph{et al.} \cite{lee2017going} explored local contextual interactions by jointly exploiting local spatio-spectral relationships of neighboring pixels. Residual learning \cite{he2016deep} was introduced by Song \emph{et al.} \cite{song2018hyperspectral} to build very deep network for extracting more discriminative feature representations for HSI classification. Li \etal \cite{li2017spectral} introduced a 3D-CNN architecture which was improved by He \etal \cite{he2017multi} to jointly learn Multi-scale 2D spatial feature and 1D spectral feature from HSI data.  Graph convolutional network (GCN) based approaches \cite{QinSTWZT19,wang2020spectral} have also been proposed which operated on graphs constructed in the \emph{spatial} domain, aggregating and transforming feature information from the neighbors of every graph node given their relevance.

Despite of the significant improvements by the DNN based methods, they suffer from a number of inherent pitfalls. Specifically, all existing methods utilize convolutional kernels with limited size of receptive field to encode the very high dimensional and informative spectral domain. As a consequence, output constituting a certain new spectral channel is derived from only a fraction of all spectral channels which blocks information flow between distant spectra and fails to capture longer term contextual information. Furthermore, spectral domain normally contains abundant yet noisy information which can lead to spurious classifications due to this weakened representation based on local contexts. 

Graphical models and graph neural networks (GNNs) have emerged as powerful tools in the field of computer vision \cite{wang2010multi,WangW14,wang2017submodular,tinghuai2020watermark,qi20173d,tinghuai2018method1,wang2019zero,zhu2019cross,zhu2019portrait,wang2019graph,tinghuai2020semantic,yang2021learning}, offering a versatile framework for representing and processing complex visual data. 
These approaches leverage the inherent structure and relationships within images \cite{wang2015robust,tinghuai2016method,tinghuai2017method,wang2020spectral,xing2021learning,han2022vision} and videos \cite{wang2010video,wang2014wide,chen2020fine}, allowing for more nuanced and context-aware analysis. By representing visual elements as nodes and their interactions as edges, graphical models and GNNs can capture spatial, temporal, and semantic dependencies that are crucial for tasks such as visual information retrieval \cite{hu2013markov}, stylization \cite{WangCSCG10,wang2011stylized,wang2013learnable}, object detection \cite{tinghuai2016apparatus,tinghuai2018method2,zhao2021graphfpn}, scene understanding \cite{deng2021generative}, and image or video segmentation \cite{wang2015weakly,wang2011probabilistic,ZhuWAK19,lu2020video,wang2021end}. In recent years, these techniques have also found significant applications in remote sensing and geospatial data analysis. For instance, in the domain of Global Navigation Satellite Systems (GNSS), graphical models have been employed to enhance positioning accuracy by modeling the intricate relationships between satellite signals, atmospheric conditions, and ground-based receivers \cite{jiang2022implementation,jiang2022smartphone}.

In this paper, we propose a novel spectral graph reasoning network (SGR) to explicitly address the above issues. Specifically, we firstly unpack and cast multiple spectral embeddings into a unified graph with each node corresponding to an individual spectral embedding channel via a novel spectral decoupling module. Graph reasoning facilitates the interactions between spectra which in turn boosts the representational capacity of the network by discriminating the significance of different spectral bands.    A spectral-specific graph embedding hierarchy is generated through the spectral decoupling module which encodes a hierarchical representation of the spectral information. Each level of the hierarchy carries different degrees of discriminative power for HSI classification. To the best of our knowledge, this is the first method to model and interpret HSI data using graph neural network in the spectral domain.

We further introduce a spectral ensembling module that correlates the multi-contextual spectral-specific graph embeddings across the spectral hierarchy. This is achieved through a novel recurrent graph propagation mechanism which learns to map from a sequence of input graphs with varying size, each representing a state of transformed spectra with its intrinsic structure, to an ensembled spectral representation containing the key discriminative features from each contextual level to improve the classification accuracy.

\section{Method}

We propose a novel end-to-end spectral-specific graph representation learning framework for HSI classification, which comprises two core modules \ie spectral decoupling module and spectral ensembling module, as illustrated in Fig. \ref{diagram}. 

\subsection{Graph Construction}

We define a graph structure for mining correlations and constraints among spectral embedding channels. Specifically, we define an undirected graph  $\mathcal{G} = (\mathcal{V}, \mathcal{E}, \mathcal{A})$. Given the feature map $X \in \mathbb{R}^{N\times H \times W}$ produced by the encoder network from an input HSI data volume centered at each pixel location, we seek to construct graph node set $\mathcal{V}=\{v_i\}_{i=1}^{N}$ and each node $v_i\in \mathcal{V}$ corresponds to a spectral feature from the embedding and the associated feature vector with $C = H\times W$ dimensions indicates $x_i\in \mathbb{R}^{C}$. The set of edges  $\mathcal{E} = \{e_{ij}\}$ indicate the connection between nodes, and $\mathbf{A} = (e_{ij})_{N\times N}$ is the adjacency matrix, with $a_{i,j}=0$ if $(i,j) \notin \mathcal{E}$, and $a_{i,j}=1$ if $(i,j) \in \mathcal{E}$. We adopt K-Nearest Neighbor Graph (KNN-Graph) construction \cite{wang2014graph,wang2016semi,wang2016primary,wang2017cross} in which two vertices $v_i$ and $v_j$ are connected by an edge, if the distance between $v_i$ and $v_j$ is among the k-th smallest distances measured by cosine distance. The normalized graph Laplacian \cite{chung1997spectral,WangW16} is computed as $L=I_N - D^{-\frac{1}{2}}AD^{-\frac{1}{2}}$, where $D^{ii}= \sum_j A^{ij}$. 

The spectral convolutions on graphs \cite{defferrard2016convolutional} can be formulated as the multiplication of input $x\in \mathbb{R}^n$ with operator $g_{\theta}$ in Fourier domain,
\begin{align}
y = g_{\theta} (L) * x = U g_{\theta} (\Lambda) U^T x
\label{eq:1}
\end{align}
where graph Fourier basis $U$ is the matrix of the eigenvectors of the normalized graph Laplacian $L$ such that $L=U\theta (\Lambda) U^T$ with $\Lambda$ being its corresponding eigenvalues, and $U^T x$ represents the graph Fourier transform of $x$. To reduce the computational complexity,  Kipf and Welling  \cite{KipfW17} stacked multiple localized graph convolutional layers with the first-order approximation of graph Laplacian,
\begin{align}
Y = \Tilde{D}^{-\frac{1}{2}}\Tilde{A}\Tilde{D}^{-\frac{1}{2}} X W
\label{eq:2}
\end{align}
where $W \in \mathbb{R}^{M\times F}$ is the matrix of filter parameters, $\Tilde{A}$ and $\Tilde{D}$ are the normalized versions with $\Tilde{A} = A + I_N$ and $\Tilde{D}^{ii}= \sum_j \Tilde{A}^{ij}$.

%Defferrard \etal{} \cite{defferrard2016convolutional} proposed using Chebyshev polynomials with K-th order to approximate filter $g_{\theta}$,
%\begin{align}
%\mathbf{y} = \sum_{k=0}^{K-1} \theta_k T_k(\Tilde{L}) \mathbf{x},
%\label{eq:2}
%\end{align}
%where $\theta \in \mathbb{R}^{K}$ is a vector of Chebyshev coefficients, and $T_k(\Tilde{L}) \in \mathbb{R}^{N\times N}$ represents the Chebyshev polynomial of order $k$ evaluated at the scaled Laplacian $\Tilde{L} = 2L/\lambda_{\text{max}} - I_n$. The Chebyshev polynomial may be computed by the stable recurrence relation $T_k(\Tilde{L}) = 2\Tilde{L} T_{k-1}(\Tilde{L}) -  T_{k-2}(\Tilde{L})$ with $T_0=1$ and $T-1=\Tilde{L}$. 

\subsection{Spectral Decoupling}

The spectral decoupling module aims to learn spectral-specific feature representation by reasoning within spectral embedding at multiple scales. Necessitated by the excessive information of the broad and noisy spectral domain, this module explores the long-range contextual dependencies in the original feature space to form multiple lower dimensional intermediate representations in order to capture the essence of the spectral information. 

We adopt a graph pyramid of spectral features - at each scale of the graph pyramid, graph convolution (GCN) is applied to capture the interdependence of graph via message passing between the nodes of graphs, \ie spectral features, as illustrated in Fig. \ref{diagram}. Assuming that filters in later layers of an encoder are responsive to various materials that make up the scanned scene, the proposed band GCN captures correlations and occurrences between more abstract characteristics in the scene like constituent materials. Graph pooling is applied to reduce the resolution of spectral features, whose result forms the input of the following graph convolution layer. 

%\subsubsection{Graph Pooling}

Graph pooling aims to enable down-sampling on graph data while preserving as much information as possible from the original graph. To this end, we employ a single neuron  linear layer parameterized by $W_{\theta}$, followed by a sigmoid function and top-$k$ pooling. This is implemented as  multiplying the input at $l$th layer  $X^l \in \mathbb{R}^{N\times C}$ with $W_{\theta}^l\in \mathbb{R}^C$ followed by sigmoid operation, whose result is an 1D vector $\mathbf{s} \in \mathbb{R}^N$ measuring how much information of nodes can be retained when projected onto the direction $W_{\theta}^l$. Graph coarsening can thus be done by selecting nodes with the largest scalar projection values on $W_{\theta}^l$ to form a new graph. The location of selected nodes in the original graph is recorded for graph unpooling operation. The feature of the selected nodes is adjusted by its projection values as $X^{l+1} = X^{l} \odot \mathbf{s}$ where $\odot$ is the Hadamard product. Intuitively the selected nodes represent the ``cluster centers'' of spectral features and the edges measure the similarity between these clusters.

%Notably, the spectral decoupling module casts multiple spectral embeddings into a unified graph, with each node of the graph being an individual spectral band in the embedding space; the graph performs interpretable reasoning to aggregate and align spectral information to guide learning spectral-specific graph embeddings at multiple contextual levels. 

\begin{figure*}[t]
	\centerline{\includegraphics[width=1.0\linewidth]{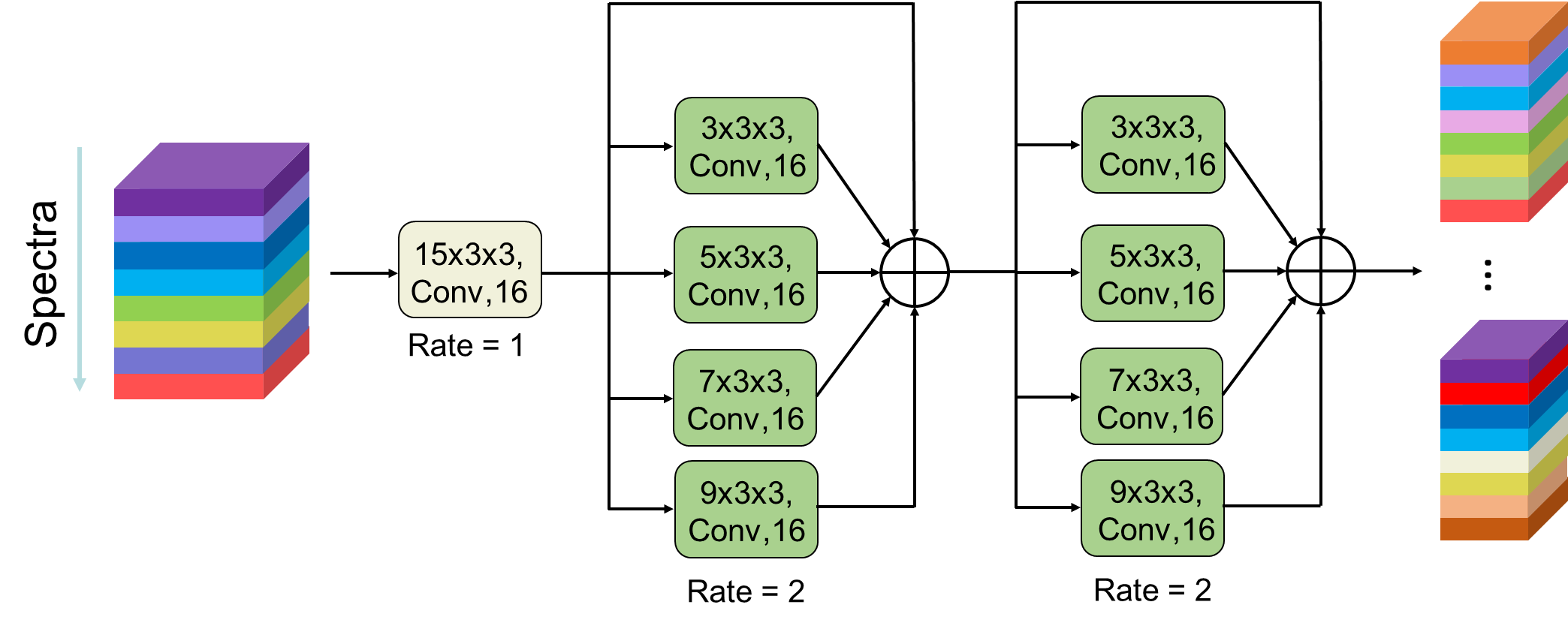}}
	\caption{Illustration of the encoder network.}
	\label{encoder}
\end{figure*}

\subsection{Spectral Ensembling}
Once obtaining a pyramid of graph embeddings after graph pooling, \ie $\mathcal{X}=\{X_1, X_2, \cdots, X_L\}$, with each level of the hierarchy encoding the key spectral-specific feature at a certain scale, we  explore their interactions across the hierarchy via a novel recurrent spectral ensembling module. There are mainly two challenges in ensembling a sequence of multi-level graph embeddings. Firstly, interaction between graphs of different number of nodes is not as straightforward as grid-like data \eg images and texts due to the lack of spatial locality and order information. Note, the number of nodes in the graph corresponds to the cardinality of spectral embeddings.  Secondly, recall that each of the previous graph pooling layer selected some important nodes to form a new graph for high-level feature encoding, the nodes in the new graph might become isolated due to the related edges have been removed while removing nodes. 

In order to address the above issues, we propose a recurrent message passing ensembling mechanism. Specifically, for every two neighboring graph embeddings, the smaller but higher level graph is firstly unpooled to match the size of the bigger yet lower level graph. Then two graphs are interacting in a way that higher level graph is used as the hidden state encoding historical information, transferring its graph embedding and node interdependency to guide the message passing within the lower level graph. Fig. \ref{diagram} illustrates the unrolled recurrent spectral ensembling where all the intermediate graph representations are processed as a sequence of input instances in the recurrent unit with shared weights. 

Formally, indicating $\mathbf{h}_{0} = X_L$, the graph unpooling layer $l$ is defined as
\begin{align}
\mathbf{\breve{h}}_{l} = \text{MessagePassing} (f (0_{N\times C}, \mathbf{h}_{l}, \mathcal{I}), \mathcal{G}_{L-l})
\label{eq:3}
\end{align}
where $\mathcal{I}$ represents the set of indices of selected nodes in the corresponding graph pooling layer $L-l$ which reduced the number of graph nodes from N to k (due to top-$k$ pooling),  $\mathbf{h}_{l} \in \mathbb{R}^{k\times C}$ is the feature map of current graph,  $0_{N\times C}$ is the empty feature matrix for the new graph, $f(\cdot)$ indicates the mapping operation distributing row vectors of $\mathbf{h}_{l}$ into $0_{N\times C}$ according to the corresponding indices in $\mathcal{I}$, $\mathcal{G}_{L-l}$ indicates the graph definition before graph pooling layer $L-l$ which consists of affinity matrix and edge weights, and $\text{MessagePassing}$ defines the message passing process \cite{KipfW17} as 
\begin{align}
\mathbf{\breve{h}}^i_{l} = \sum_{j \in \mathcal{N}(i) \cup \{ i \}} \mathbf{\Theta} \cdot \mathbf{h}^j_{l},
\label{eq:4}
\end{align}
where neighboring node features are first transformed by the weight matrix $\mathbf{\Theta}$ passed via $\mathcal{G}_{L-l}$. The potential isolation of upsampled nodes in the new graph due to the removed related edges is properly compensated by the message passing based on the original graph interdependency information. 

Then the framework updates the hidden state based on spectral feature $X_{L-l}$ and  hidden state $\mathbf{h}_{l-1}$ from previous step via a gated mechanism similar to Gated Recurrent Unit (GRU) \cite{ChoMGBBSB14} as
\begin{equation}\label{eq:GRU}
\begin{split}
    \mathbf{z}_{l} &= \sigma \left(\mathbf{W}_z  X_{L-l} +
        \mathbf{U}_z  \mathbf{\breve{h}}_{l-1}\right), \\
    \mathbf{r}_{l} &= \sigma \left(\mathbf{W}_r  X_{L-l} +
        \mathbf{U}_r  \mathbf{\breve{h}}_{l-1}\right), \\
    \mathbf{\tilde h}_{l} &= \text{tanh} \left(
        \mathbf{W}_o  X_{L-l} + \mathbf{U}_o (\mathbf{r}_{l} \odot \mathbf{\breve{h}}_{l-1})\right), \\
    \mathbf{h}_{l} &= (1-\mathbf{z}_{l}) \odot \mathbf{\breve{h}}_{l-1} +
        \mathbf{z}_{l} \odot \mathbf{\tilde h}_{l}
\end{split}
\end{equation}
where $\sigma(\cdot)$ is the logistic sigmoid function, $\text{tanh}(\cdot)$ is the hyperbolic tangent function, and $\odot$ is the Hadamard product. Interactions between all levels in the graph embedding hierarchy are enabled by encoding higher level contextual spectral information in the hidden states. The process is repeated L times according to the levels of hierarchy, and the final hidden state $\mathbf{h}_{L}$ goes through another GCN layer before being concatenated with the feature map from the encoder network for predicting the category label of the input HSI volume with a fully connected layer.

\subsection{Encoder Network}
We adopt a multi-path dilated 3D convolutional residual network to encode the orginal HSI data as illustrated in Fig. \ref{encoder}. Specifically, the encoder starts with a 3D convolutional layer with kernel size $15\times 3 \times 3$ corresponding to the spectral dimension and two spatial dimensions respectively, which is followed by two multi-path ($3\times 3 \times 3$, $5\times 3 \times 3$, $7\times 3 \times 3$ and $9\times 3 \times 3$ kernels respectively) dilated (rate=2) convolutional residual modules with identity shortcuts. Note that our proposed framework is agnostic to the utilized encoder network. 

\begin{table*}[!t]
	\centering
	%\begin{center}
	\begin{tabular}{clcccccc}
		\hline
		Class No.  & Class Name  & Training & Testing \\
		\hline
		1&Asphalt & 50 & 6581 \\
		2&Meadows & 50 & 18599 \\
		3&Gravel & 50 & 2049 \\
		4&Trees & 50 & 3014\\
		5&Painted metal sheets & 50 & 1295 \\
		6&Bare Soil & 50 & 4979 \\
		7&Bitumen & 50 & 1280 \\
		8&Self-Blocking Bricks & 50 & 3632 \\
		9&Shadows & 50 & 897 \\
		\hline
		&Total & 450 & 42776\\
		\hline
		%\vfill
		%\toprule[1pt]
	\end{tabular}
	%\end{center}
	\caption{Sample size for the University of Pavia dataset.}%and \textbf{IS}: deep supervision
	\label{table:pu-sample}
\end{table*}

\begin{table*}[!t]
	\centering
	%\begin{center}
	
	\begin{tabular}{clccccccc}
		\hline
		Class No.  & Class Name & Training & Testing \\
		\hline
		1&Alfalfa & 40 & 6 \\
		2&Corn-notill& 100 & 1328\\
		3&Corn-mintill& 100 & 730\\
		4&Corn& 100  & 137\\
		5&Grass-pasture& 100 &383\\
		6&Grass-trees& 100& 630\\
		7&Grass-pasture-mowed& 20  & 8\\
		8&Hay-windrowed& 100& 378\\
		9&Oats& 15 & 5\\
		10&Soybean-notill&  100& 872\\
		11&Soybean-mintill&  100& 2355\\
		12&Soybean-clean& 100 & 493\\
		13&Wheat& 100 & 105\\
		14&Woods& 100 & 1165\\
		15&Buildings-Grass-Trees-Drives& 100 & 286\\
		16&Stone-Steel-Towers& 80 & 13\\
		\hline
		&Total & 1355 & 8894\\
		\hline
		%\vfill
		%\toprule[1pt]
	\end{tabular}
	%\end{center}
	\caption{Sample size for the Indian Pines dataset}%and \textbf{IS}: deep supervision
	\label{table:ip-sample}
\end{table*}

\section{Experimental Results}
We evaluate our method on two publicly available hyperspectral image datasets with four metrics including per-class accuracy, overall accuracy (OA), average accuracy (AA), and Kappa coefficient. The network architecture of the proposed method SGR is identical for both datasets. Specifically, we use an architecture of L=2 and the input HSI is sampled on 7$\times$7 image patches in the spatial domain.  We adopt stochastic gradient descent with momentum set to 0.9, weight decay of 0.0005, and with a batch size of 30. We initialize an equal learning rate for all trainable layers to 0.05, which is manually decreased by a factor of 10 when the validation error plateaus.  The number of training epochs is set to 500. All the reported accuracies are calculated based on the average of five training sessions to obtain stable results.

\subsection{Datasets}

The University of Pavia dataset was captured with the ROSIS sensor in 2001 which consists of 610$\times$340 pixels with a spatial resolution of 1.3 m$\times$1.3 m and has 103 spectral channels in the wavelength range from 0.43 $\mu$m to 0.86 $\mu$m after removing noisy bands. This dataset includes 9 land-cover classes and the false color image and ground-truth map are shown in Fig. \ref{fig:pu-result}.

The Indian Pines dataset was captured with Airborne Visible/Infrared Imaging Spectrometer sensor comprising 145$\times$145 pixels with a spatial resolution of 20 m$\times$20 m and 220 spectral channels covering the range from 0.4 $\mu$m to 2.5 $\mu$m.  This dataset includes 16 land-cover classes and the false color image and ground-truth map are shown in Fig. \ref{fig:indianpines-result}.

Table \ref{table:pu-sample} and Table \ref{table:ip-sample} provide information about all the classes of both datasets with their corresponding training and test samples respectively. 
During training, 90\% of the training samples are used to learn the network parameters and 10\% are used as validation set for tuning the hyperparameters.

\iffalse
The Kennedy Space Center dataset which was taken by AVIRIS sensor over Florida with a spectral coverage ranging from 0.4 $\mu$m to 2.5 $\mu$m, contains 224 bands and 614 $\times$ 512 pixels with a spatial resolution of 18 m. This dataset comprises 13 land-cover classes as listed in Table \ref{table:ksc}, and Fig. \ref{fig:ksc-result} exhibits the false color image and ground-truth map.
\fi

\subsection{Classification Results}

\begin{table*}[!t]
	\centering
	%\begin{center}
	\begin{tabular}{cccccccccc}
		\hline
		Class No. & FCN \cite{lee2016contextual} & MS 3D-CNN \cite{he2017multi}  & SS 3D-CNN \cite{li2017spectral} & 3D-CNN \cite{hamida20183} & SGR \\
		\hline
		1 & 61.98 & 92.47 & 91.98 & 91.86 & \textbf{95.91} \\
		2 & 72.16 & 94.22 & 93.88 & 90.55 & \textbf{98.27} \\
		3 & 12.90 & 86.38 & 81.46 & 85.89 & \textbf{88.47} \\
		4 & 39.52 & 92.63 & 94.98 & 93.65 & \textbf{99.83} \\
		5 & 99.46 & 98.57 & \textbf{100.00} & 98.59 & 99.91 \\
		6 & 41.78 & 80.45 & 74.34 & 70.49 & \textbf{84.75} \\
		7 & 50.49 & 83.60 & 84.94  & 84.66 & \textbf{86.42} \\
		8 & 67.99 & 90.30 & 86.92 & 88.58 & \textbf{91.43}\\
		9 & 97.55 & 97.43 & \textbf{99.94} & 98.82 & 98.24 \\
		\hline
		OA & 61.21 & 91.35  & 90.30  & 88.14 & \textbf{94.37}\\
		AA & 60.42 & 90.67 & 89.82 & 89.23 & \textbf{93.69}\\
		Kappa & 48.44 & 88.56 & 87.15 & 84.46 & \textbf{91.26}\\
		\hline
		%\vfill
		%\toprule[1pt]
	\end{tabular}
	%\end{center}
	\caption{Per-class accuracy, OA, AA (\%), and Kappa coefficient achieved by different methods on the University of Pavia dataset.}%and \textbf{IS}: deep supervision
	\label{table:pu}
\end{table*}

\subsubsection{The University of Pavia Dataset}
The quantitative results obtained by different approaches on the University of Pavia dataset are reported in Table \ref{table:pu}. Our method outperforms the compared methods on 7 out of 9 categories and achieves the best overall results, \ie  OA, AA and Kappa. Generally, all other networks significantly outperform  FCN \cite{lee2016contextual}. One possible reason could be that FCN heavily squeezed the spectral dimension and utilized small 2D convolutional kernels in the spatial domain. 3D-CNN \cite{hamida20183} adopted more 3D convolutional layers to perform simultaneous spatial-spectral convolution. Compared with \cite{lee2016contextual}, \cite{hamida20183} used a large number of 3D filters per layer to improve the feature representation over the very high dimensional spectral domain which increases the accuracy by $\sim 2\%$. \cite{he2017multi} addresses the limitations of single scale 3D CNN architectures by introducing a multi-scale 3D CNN architecture, which improves the accuracy by $\sim 1\%$ compared to \cite{hamida20183}. Nevertheless, none of the existing methods has explicitly performed reasoning among the spectral embedding channels in a global manner, but rather modeling the local context by convolutional kernels with limited size of receptive field. Our method bridges this gap by  
casting spectral embeddings into a unified graph whose node corresponds to an individual spectral feature in the embedding space and thus the interaction between graph nodes guides learning spectral-specific graph representations. Quantitative results shows that our method outperforms the best competing method \cite{he2017multi} with a significant margin of $3.02\%$.

\begin{figure*}[!ht]
	\centering
	%\begin{center}
	%\captionsetup[subfigure]{justification=centering}
	\begin{subfigure}{0.23\textwidth}
		\includegraphics[height=5cm,width=0.99\linewidth]{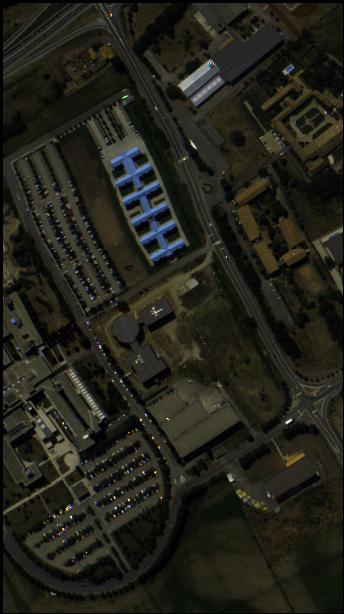}
		%\caption{(a)}
		\label{fig:Image}
	\end{subfigure}%
	%\hfill %%useful if width of each figure is less the 
	\begin{subfigure}{0.23\textwidth}
		\includegraphics[height=5cm,width=0.99\linewidth]{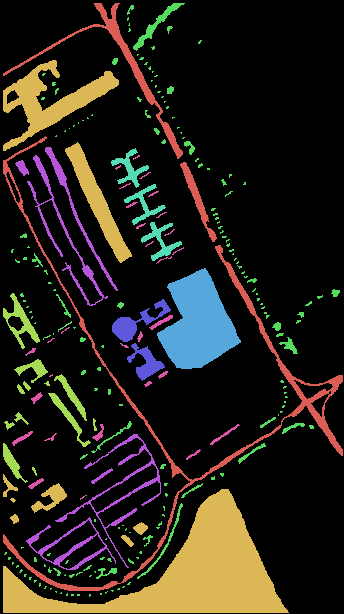}
		%\caption{(b)}
		\label{fig:Deeplab_v3_plus_16}
	\end{subfigure}%
	%\hfill % <-- added
	\begin{subfigure}{0.23\textwidth}
		\includegraphics[height=5cm,width=0.99\linewidth]{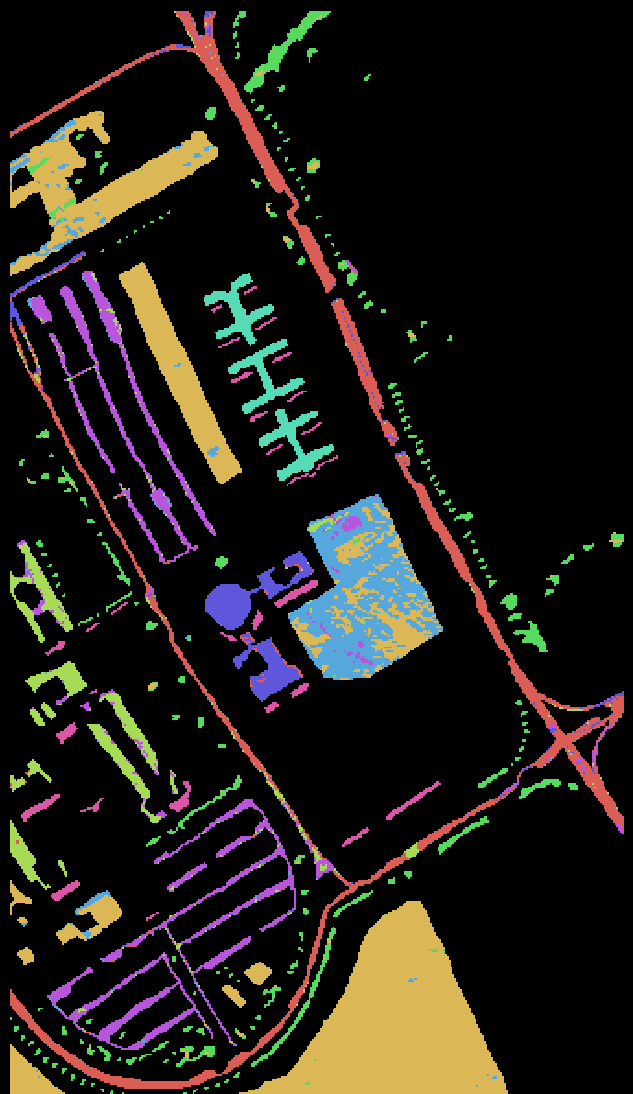}
		%\caption{(c)}
		\label{fig:MeshNet}
	\end{subfigure}
	\begin{subfigure}{0.23\textwidth}
		\includegraphics[height=5cm,width=0.99\linewidth]{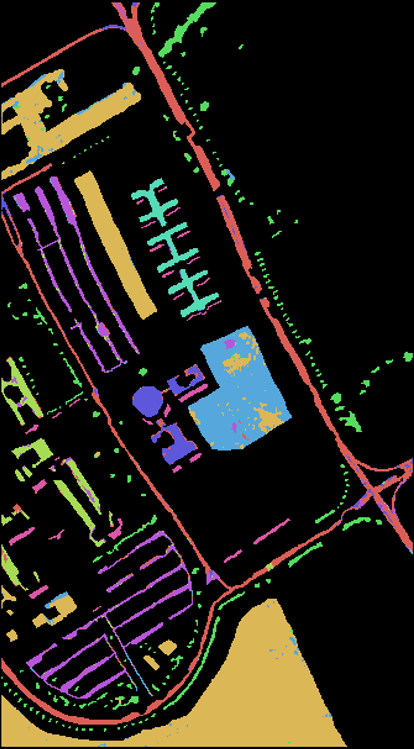}
		%\caption{(c)}
		\label{fig:MeshNet}
	\end{subfigure}
	\caption{The university of Pavia dataset: (a) False color image (b) Groundtruth map (c) Prediction of SSLSTMs \cite{zhou2019} (d) Prediction of the proposed SPGAT}
	\label{fig:pu-result}
\end{figure*}

Fig. \ref{fig:pu-result} shows a visual comparison of our proposed SGR and the best competing method \cite{he2017multi} on the University of Pavia dataset. The observation reveals that \cite{he2017multi} suffers more mis-classifications than SGR in large semantically coherent regions, due to its weaker representation capability of encoding the discriminative spectral features with the presence of noise. The qualitative result that our SGR produces more accurate and coherent predictions.

\begin{table*}[!t]
	\centering
	%\begin{center}
	\begin{tabular}{lccccccc}
		\hline
		Class No. & FCN \cite{lee2016contextual} & SS 3D-CNN \cite{li2017spectral} & MS 3D-CNN \cite{he2017multi} & 3D-CNN \cite{hamida20183} & SGR \\
		\hline
		1 & 6.7 & 80.00 & \textbf{100.00} & 30.77 & 99.56 \\
		2& 64.81 & 76.41 & 77.52 & 70.90 & \textbf{79.32} \\
		3& 74.67 & 78.06 & 79.36 & 76.12 & \textbf{80.47} \\
		4& 59.21 & 76.65 & 80.85 & 77.46 & \textbf{83.72} \\
		5& 92.19 & 92.16 & 96.43 & 90.96 & \textbf{97.62} \\
		6& 97.66 & 99.13 & 99.84 & 96.86 & \textbf{99.92} \\
		7& 47.56 & 40.00 & \textbf{100.00}& 48.48 &  \textbf{100.00} \\
		8& 97.04 & 99.87 & \textbf{100.00} & 97.70 & \textbf{100.0} \\
		9& 16.95 & 58.82 & \textbf{100.00} & 71.43 & \textbf{100.00} \\
		10& 78.71 & 76.75 & 79.84 & 70.07 & \textbf{81.55} \\
		11& 76.59 & 80.09 & 82.50 & 77.83 & \textbf{85.26} \\
		12& 65.23 & 78.03 & 76.56 & 68.31 & \textbf{80.97} \\
		13& 98.51 & 97.65 & 99.53 & 95.57 & \textbf{99.92} \\
		14& 94.88 & 94.64 & 97.34 & 96.12 & \textbf{98.75} \\
		15& 70.75 & 73.96 & 85.53 & 71.70 & \textbf{85.97}\\
		16& 49.33 & 89.66 & \textbf{100.00} & 78.79 & \textbf{100.00}\\
		\hline
		OA & 79.20 & 83.37 & 85.79 & 80.35 & \textbf{87.83}\\
		AA & 68.17 & 80.74 & 90.96 & 76.19 & \textbf{92.06}\\
		Kappa & 75.89 & 80.89 & 83.58 & 77.19 & \textbf{85.42}\\
		\hline
		%\vfill
		%\toprule[1pt]
	\end{tabular}
	%\end{center}
	\caption{Per-class accuracy, OA, AA (\%), and Kappa coefficient achieved by different methods on the Indian Pines dataset}%and \textbf{IS}: deep supervision
	\label{table:ip}
\end{table*}

\subsubsection{The Indian Pines Dataset}

\begin{figure*}[!ht]
	\centering
	%\begin{center}
	%\captionsetup[subfigure]{justification=centering}
	\begin{subfigure}{0.23\textwidth}
		\includegraphics[height=5cm,width=0.99\linewidth]{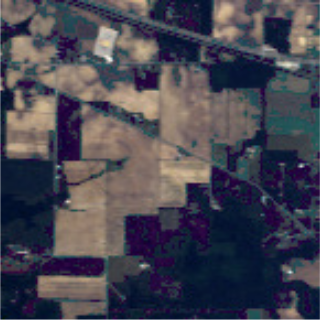}
		%\caption{(a)}
		\label{fig:Image}
	\end{subfigure}%
	%\hfill %%useful if width of each figure is less the 
	\begin{subfigure}{0.23\textwidth}
		\includegraphics[height=5cm,width=0.99\linewidth]{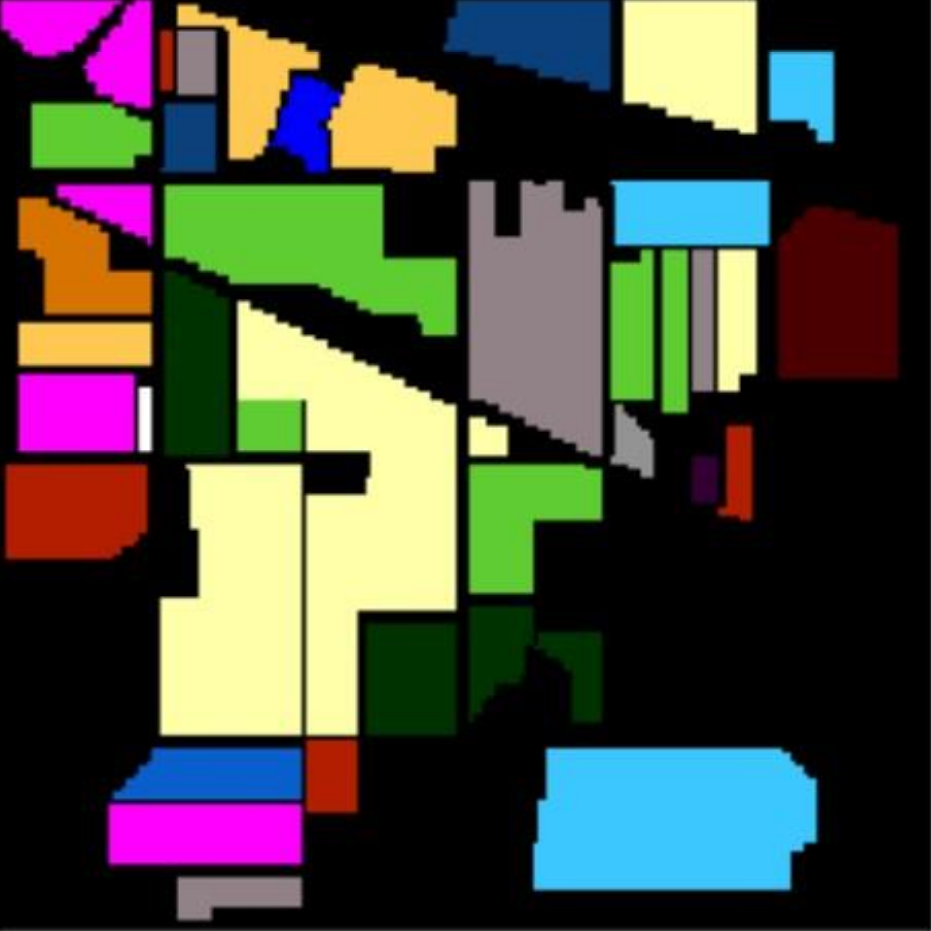}
		%\caption{(b)}
		\label{fig:Deeplab_v3_plus_16}
	\end{subfigure}%
	%\hfill % <-- added
	\begin{subfigure}{0.23\textwidth}
		\includegraphics[height=5cm,width=0.99\linewidth]{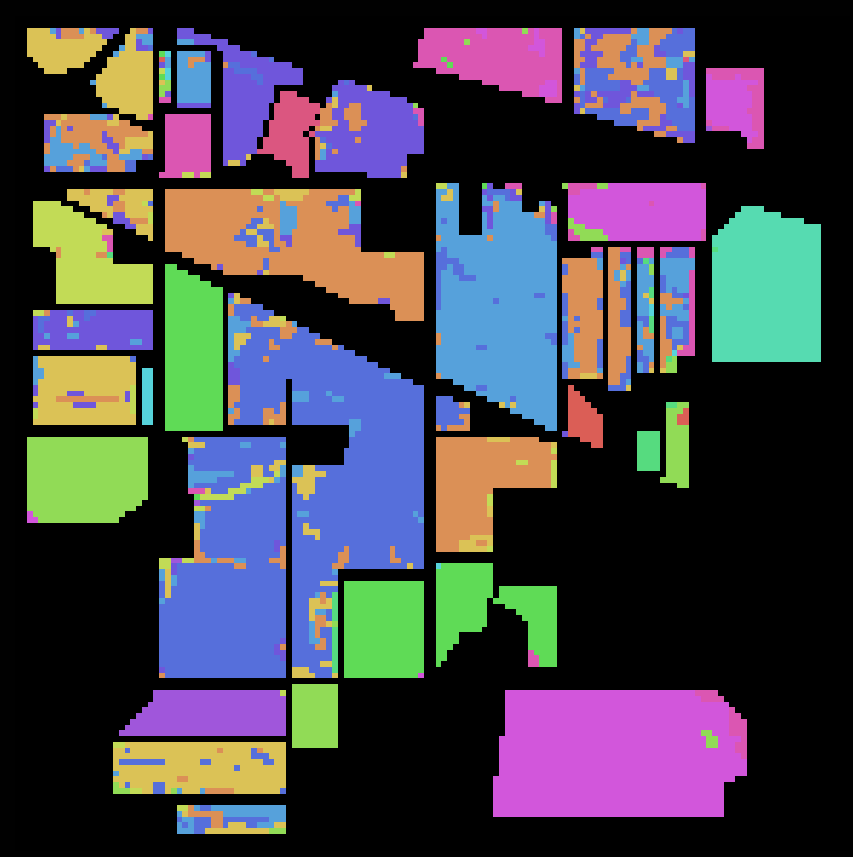}
		%\caption{(c)}
		\label{fig:MeshNet}
	\end{subfigure}
	\begin{subfigure}{0.23\textwidth}
		\includegraphics[height=5cm,width=0.99\linewidth]{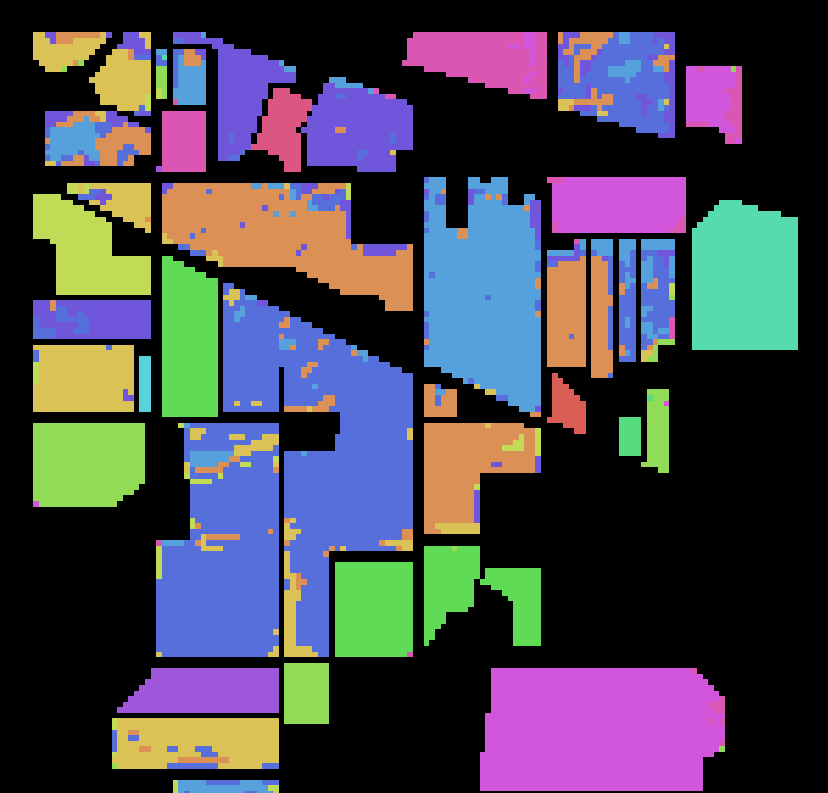}
		%\caption{(c)}
		\label{fig:MeshNet}
	\end{subfigure}
	\caption{Indian Pines dataset: (a) False color image (b) Groundtruth map (c) Prediction of LBMSELM \cite{CaoYRCHS19} (d) Prediction of the proposed SPGAT}.
	\label{fig:indianpines-result}
\end{figure*}

Table \ref{table:ip} gives the quantitative results on the Indian Pines dataset, which shows that the proposed method obtains the best overall accuracy of $87.83\%$. 
Similar observations with the University of Pavia dataset can be found that all 3D CNN architectures outperform FCN  \cite{lee2016contextual} which adopts small 2D convolutional kernels as majority components. 3D CNN architectures generally explore the spatial-spectral contexts simultaneously and achieves better accuracy. Nevertheless, our proposed method has the advantage over the compared methods by exploring inter-spectral relations and learning a graph embedding hierarchy representation with each level emphasizing on specific spectral features exhibiting discriminative power toward the classification. Fig. \ref{fig:indianpines-result} shows a visual comparison of our proposed SGR and the second best method \cite{he2017multi}, which confirms that our SGR produces more accurate and coherent predictions.

\subsection{Ablation Study}

We conduct ablation study on the University of Pavia dataset to quantitatively verify the effectiveness of the proposed architecture. The first baseline is 
applying a predicting layer (Conv3D-FC-Softmax) on top of the encoder network, \ie Baseline-1. We also investigate spectral decoupling module with two different depth 1 and 3, dubbed as Baseline-2 and Baseline-3 respectively. To verify the contribution from spectral ensembling, we replace it with a naive graph upsampling and embedding summation operation, \ie Baseline-4.  

As shown in Table \ref{table:ablation}, a significant accuracy gain of $6.13\%$ is achieved by our proposed spectral graph reasoning network, by comparing Baseline-1 and SGR. Adding one level of spectral coupling module improves the accuracy from $88.24\%$ to $91.87\%$, which confirms its effectiveness. However, adding  three levels slightly decreases the accuracy, \ie a loss of $0.16\%$ compared with SGR which has two levels. The possible reason is using the limited training samples for learning a more complex architecture. Furthermore, introducing the spectral ensembling module boosts the performance by $1.83\%$.

%One challenge in hyperspectral data classification is that due to complex light scattering mechanism, some pixels of a hyperspectral image, which belong to the same land cover class, have different spectral signatures. Therefore, an approach that is capable of making spectral signals of those pixels that are more similar should be able to offer a more accurate classification result.

\begin{table*}[!t]
	\centering
	%\begin{center}
	\begin{tabular}{lccccccc}
		\hline
		& Baseline-1 & Baseline-2 & Baseline-3 & Baseline-4 & SGR \\
		\hline
		OA & 88.24 & 91.87 & 94.21 & 92.54 & 94.37\\
		%OA & 88.24 & 92.22 & 94.21 & 93.24 & 94.37\\
		\hline
		%\vfill
		%\toprule[1pt]
	\end{tabular}
	%\end{center}
	\caption{OA achieved by different baselines on the University of Pavia dataset.}%and \textbf{IS}: deep supervision
	\label{table:ablation}
\end{table*}

\section{Conclusions}
In this paper, we proposed a novel spectral graph reasoning network for hyperspectral image classification which has significantly advanced the state-of-the-art. At the core of our network architecture lies a new spectral feature reasoning and ensembling paradigm. We demonstrated that applying interpretable graph reasoning in the spectral feature domain enables learning a spectral-specific graph embedding which in turn improves the discriminative capability.  We further proposed a  spectral ensembling module that explores the interactions and interdependencies across the graph embedding hierarchy via a novel recurrent message propagation mechanism.

% EITHER use the included BST file
\bibliographystyle{splncs03}
\bibliography{egbib}
\end{document}